\documentclass[journal]{IEEEtran}
\usepackage{cite}
\usepackage{amsmath,amssymb,amsfonts}
\usepackage{algorithmic}
\usepackage{graphicx}
\usepackage{textcomp}
\usepackage{xcolor}
\usepackage{booktabs}
\usepackage{multirow}
\usepackage{makecell}
\usepackage{hyperref}
\usepackage{tablefootnote}

\def\BibTeX{{\rm B\kern-.05em{\sc i\kern-.025em b}\kern-.08em
    T\kern-.1667em\lower.7ex\hbox{E}\kern-.125emX}}

\begin{document}

\title{Event Stream based Multi-Modal Video Anomaly Detection: A Benchmark Dataset and Algorithms}


\author{\makecell{Peipei Zhu*, Yueqing Niu*, Lin Zhu, Guanchong Niu, Yang Yu, Zheng Li}
 \thanks{
 Peipei Zhu, Yueqing Niu, and Yang Yu are with the College of Pharmaceutical Engineering of Traditional Chinese Medicine, Tianjin University of Traditional Chinese Medicine, Tianjin 301617, China (e-mail: zhupp0527@tjutcm.edu.cn, 1035415060@qq.com, and yu\_yang@tjutcm.edu.cn).

Lin Zhu is with the School of Artificial Intelligence, Beijing Normal University, Beijing 100875, China. (e-mail: linzhu@pku.edu.cn).

Guanchong Niu is with the Guangzhou Institute of Technology, Xidian University, Guangzhou 510555, China. (e-mails: niuguanchong@xidian.edu.cn).

Zheng Li is with the College of Pharmaceutical Engineering of Traditional Chinese Medicine, Tianjin University of Traditional Chinese Medicine, Tianjin 301617, China, and also with State Key Laboratory of Component-based Chinese Medicine, Tianjin University of Traditional Chinese Medicine, Tianjin 301617, China (e-mail: lizheng@tjutcm.edu.cn).

Corresponding author: Zheng Li. *The authors have contributed equally to this work.
 }
}

\maketitle

\begin{abstract}
Video anomaly detection (VAD) is critical for automated surveillance but remains fragile under challenging conditions such as illumination variations, fast motion, and complex backgrounds when relying solely on visible-light videos. To address these limitations, we propose E-VAD, an event-enhanced VAD framework that jointly exploits conventional video and event streams captured by bio-inspired event cameras. Event sensors asynchronously capture brightness changes with high temporal resolution, offering robustness to motion blur and extreme lighting, and providing motion-salient cues complementary to video-based visual information. To support multi-modal VAD research, we construct a large-scale visible–event benchmark comprising 6.3 billion events and 376,368 video frames collected under diverse illumination levels, motion patterns, and background complexities, filling the gap of realistic and scalable datasets for event-based anomaly detection. Building upon this dataset, we design a contrastive multi-modal pretraining framework to learn discriminative event representations by aligning semantic embeddings across event streams, visible videos, and textual descriptions. An adaptive fusion module then dynamically integrates event-based temporal cues with video-based spatial semantics, improving robustness to environmental disturbances. Experiments on benchmarks and the proposed TJUTCM Pha dataset demonstrate that E-VAD consistently outperforms methods, validating the effectiveness of event-based sensing for VAD in real-world scenarios.
\end{abstract}

\section{Introduction}

\IEEEPARstart
Video anomaly detection (VAD) plays a central role in intelligent surveillance, autonomous perception, and industrial safety, enabling automatic identification of irregular patterns in continuous video streams \cite{narwade2024synthetic, zhang2024multi}. As VAD systems are increasingly deployed in manufacturing \cite{ahn2023safefac, chen2025large, liu2024artificial}, transportation \cite{bogdoll2022anomaly}, and public security infrastructures \cite{ullah2023comprehensive}, they are expected to remain reliable under fluctuating illumination, rapid motion, and operational complexity. Benefiting from the growth of visible-light datasets and deep representation learning \cite{zhu2022unpaired, ren2024super, liu2024iterative, bao2023intelligent}, current VAD research has converged toward a standard pipeline composed of spatial feature extraction, temporal modeling, and anomaly scoring \cite{zhong2025two, xie2025distributed, zhong2024inter}.

\begin{figure} 
\centering 
\includegraphics[width=0.9\linewidth]{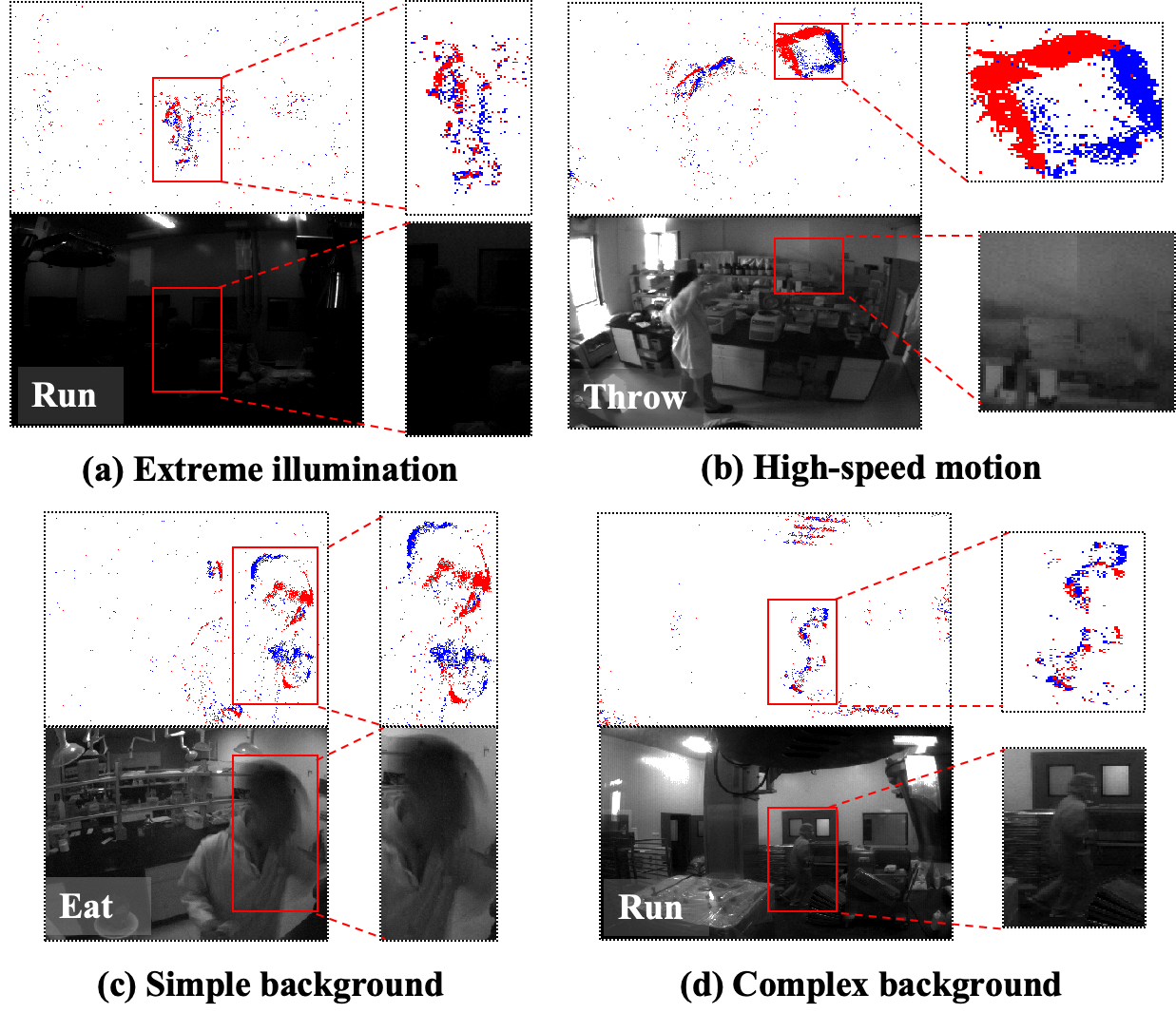} \caption{Comparison of sensing characteristics between visible-light cameras and event cameras in VAD. Subfigures (a–d) illustrate scenarios where event streams provide clearer anomaly cues than conventional video, including extreme illumination, high-speed motion, and both simple and cluttered environments. In these settings, the sparse, change-driven responses of event sensors suppress static background interference and enhance the signal-to-noise ratio of anomalous behaviors, enabling more reliable recognition of rapid or subtle irregularities. By integrating visible frames with event streams, VAD systems can simultaneously exploit spatial fidelity and temporally precise motion cues, making event information not just complementary but fundamentally indispensable for robust anomaly detection in real-world environments.} \label{fig:placeholder} \end{figure}

Despite these advances, nearly all existing benchmarks rely solely on conventional visible-light videos, whose sensing characteristics impose fundamental constraints on anomaly perception. As global-frame cameras, visible sensors degrade sharply when imaging conditions deviate from ideal visibility, such as illumination fluctuations cause contrast imbalance, fast motion introduces smear that destroys precise motion cues, and low signal-to-noise environments suppress fine behavior cues \cite{gehrig2024low}. Real monitoring examples further reveal these issues, including rapid operations induce severe blur, reflective materials lead to unstable brightness \cite{wu2024leod}, and low-light conditions collapse abnormal actions into ambiguous frames \cite{wang2024event}. Such degradations propagate directly into feature encoders and limit generalization, indicating that the obstacle is not merely algorithmic but stems from the inherent sensing and modeling limitations of video-based visible acquisition.

Motivated by these limitations, neuromorphic vision using event cameras offers a complementary sensing paradigm. Event cameras asynchronously capture pixel-wise brightness changes with microsecond resolution, suppressing static redundancy and remaining robust under extreme illumination and fast motion \cite{gallego2020event}. As illustrated in Fig. \ref{fig:placeholder}, event streams preserve high-fidelity temporal dynamics, while visible videos provide dense spatial semantics such as texture and scene layout. Effective anomaly detection should therefore exploit their complementarity rather than rely on a single modality. However, despite this clear advantage, the community still lacks a unified benchmark that jointly captures visible videos and event streams for anomaly detection, hindering systematic multi-modal research. More broadly, this work challenges the long-standing assumption of reliable video-based sensing in VAD and reframes robustness as a sensing-aware problem. By introducing real event-based sensing, VAD can access temporal dynamics fundamentally unavailable to conventional cameras, opening a new direction where robustness is driven by sensing diversity rather than model complexity alone.

To bridge this gap, TJUTCM Pha is introduced as the first real-scene visible–event dataset designed for anomaly detection in pharmaceutical environments. The dataset contains 6.3 billion event points and 376,368 video frames collected from laboratories and production sites with diverse illumination levels, motion variations, and background complexities, including 13 well-defined anomalous behavior categories. Compared with synthetic or screen converted datasets \cite{qian2025ucf}, TJUTCM Pha is recorded through real capture, ensuring true physical temporal dynamics and degradations. The dataset exposes model robustness to challenging conditions where visible videos degrade yet event streams remain informative. It therefore enables systematic research on sensing complementarity and establishes a foundation for advancing multi-modal VAD.

On top of this dataset, we introduce E-VAD, a multi-modal anomaly detection framework that unifies visible videos, event streams, and natural language supervision into a cohesive representation. At its core is a cross-modal contrastive pretraining module that aligns heterogeneous modalities and bridges the gap between video-based and event-driven data. This module enhances representations of sparse, high-speed event signals and improves generalization across diverse scenes. In addition, an adaptive fusion mechanism dynamically balances temporal event cues and spatial video features according to real-time signal reliability, mitigating noise from background clutter, motion blur, and illumination changes. To the best of our knowledge, this is the first framework that jointly exploits event temporally precise dynamic cues and video-text spatial semantics within a unified anomaly detection pipeline. Extensive experiments on public benchmarks and the new TJUTCM Pha dataset show that E-VAD achieves state-of-the-art performance, demonstrating that visible–event integration with semantic guidance is essential for robust, high-fidelity video anomaly detection and marks a clear advance over video-only or event-only approaches.

Our contributions are summarized as follows:
\begin{itemize}
\item TJUTCM Pha is released as the first real-scene dataset that jointly records visible videos and event streams for anomaly detection. To the best of current knowledge, no existing work offers a visible–event benchmark of comparable scale or realism for VAD research. 
\item A new multi-modal anomaly detection framework named E-VAD is developed, which unifies visible frames, event signals, and textual supervision within a single contrastively aligned representation space. The model further introduces adaptive fusion to calibrate temporal event cues against spatial appearance information, enabling reliable detection even under severe lighting fluctuation and fast object motion. This design advances the state of multi-modal VAD beyond conventional video-only architectures. 
\item Extensive evaluations highlight the clear superiority of the proposed method. E-VAD consistently surpasses representative state-of-the-art baselines on public benchmarks as well as on TJUTCM Pha. The empirical evidence confirms that the integration of event sensing and visible imaging is not simply beneficial but essential for next-generation anomaly detection systems.
\end{itemize}

The remainder of the paper is organized as follows: Section~\ref{sec:rela} reviews related work; Section~\ref{sec:meth} presents the proposed method; Section~\ref{sec:data} describes the dataset; Section~\ref{sec:expe} reports experimental results; and Section~\ref{sec:conc} concludes the paper.

\section{Related Work} \label{sec:rela}

\subsection{Video Anomaly Detection}

Existing VAD research follows two major directions, uni-modality methods operating solely on visible frames, and multi-modality approaches incorporating auxiliary cues \cite{flaborea2023multimodal}. Uni-modality frameworks avoid cross-modal alignment and build representations directly from visible sequences, including spatiotemporal reconstruction \cite{wang2022video}, diffusion-based reasoning \cite{liu2024vadiffusion, guo2025aligning}, multi-granularity temporal modeling \cite{zhang2024multi}, and memory-based predictions \cite{guo2023learning}. Although effective under constrained conditions, these methods remain limited by the physical properties of traditional cameras. To address these shortcomings, recent works explore auxiliary modalities such as textual descriptions \cite{yang2024text}, human skeleton trajectories \cite{flaborea2023multimodal}, and object-level cues \cite{georgescu2021background, georgescu2021anomaly}, further strengthened by advances in vision–language alignment \cite{wu2024open}. However, because these modalities are all extracted from the same visible stream, they inevitably inherit its limitations. As a result, even with richer semantic supervision, such approaches cannot overcome the core bottlenecks of visible-only VAD, including limited temporal sensitivity and vulnerability to illumination variation.

The proposed work establishes a new perspective beyond this paradigm. We introduce asynchronous event signals as an independent sensory stream rather than a derivative extension of video. Event data is inherently responsive to motion and brightness variation at microsecond resolution. The integration of event and video information forms a detection mechanism that does not merely enhance video-based models but redefines the sensing foundation of anomaly understanding.

\subsection{Event-based Vision}

Event cameras operate on asynchronous pixel triggered sensing which encodes brightness changes with microsecond precision. This imaging principle achieves low latency perception and maintains wide dynamic range, low power consumption and robustness under challenging conditions \cite{gehrig2024low}. Consequently, event-based perception has rapidly advanced in tasks including object detection \cite{wu2024leod, wang2025object}, action reasoning \cite{chakravarthi2024recent}, and neural radiance field reconstruction under adverse environments \cite{feng2025ae}. Previous work proposes self-supervised pre-training tailored for event signatures \cite{yang2023event}, self-training detections \cite{wu2024leod}, state-space structured event perception \cite{zubic2024state}, and hybrid radiance field learning for dynamic scenes \cite{feng2025ae}.

Although these methods demonstrate the capability of event sensing, the focus largely remains within event-based recognition or perception tasks that do not directly address anomaly detection. Furthermore, the majority of works do not exploit synergy between event signals and visible semantic structure. In contrast, our framework introduces adaptive fusion where event motion salience and visible spatial context reinforce each other. This dual stream mechanism mitigates motion blur, exposure failure and low contrast conditions that limit video-based anomaly detection. The resulting representation is inherently more resilient and better aligned with real-world safety monitoring requirements.

\subsection{Vision-Language Models}

Vision-language models have reshaped multi-modal learning by coupling high-level semantics with visual embeddings \cite{zhu2023prompt}. Contrastive training on large-scale image-text collections enables generalization across recognition tasks \cite{radford2021learning}. Subsequent research extends this paradigm to multiple heterogeneous input domains, including audio-vision alignment \cite{chen2023vlp}, point cloud-text fusion \cite{zhang2022pointclip}, temporal video-text synchronization \cite{wasim2023vita}, multi-depth perception \cite{zhang2022can}, and emerging event-text alignment frameworks \cite{wu2023eventclip, zhou2024eventbind}. For example, EventCLIP aligns events with text using grayscale event frames as visual intermediaries \cite{wu2023eventclip}. EventBind further constructs a unified embedding space for \textbf{images}, event streams and texts \cite{zhou2024eventbind}.

Drawing inspiration from these developments, our method introduces cross-modality representation learning specifically for anomaly detection. Instead of treating text merely as annotation, we utilize linguistic supervision to guide semantic alignment between visible \textbf{videos} and event streams. This creates a unified embedding space where motion-dependent event signatures and object-level video cues are jointly grounded in language semantics. The resulting model improves sensitivity to subtle and rare anomalies and demonstrates a level of robustness unattainable through video only or text augmented video systems.

\section{Methodology}\label{sec:meth}

\begin{figure*}
    \centering
    \includegraphics[width=1\linewidth]{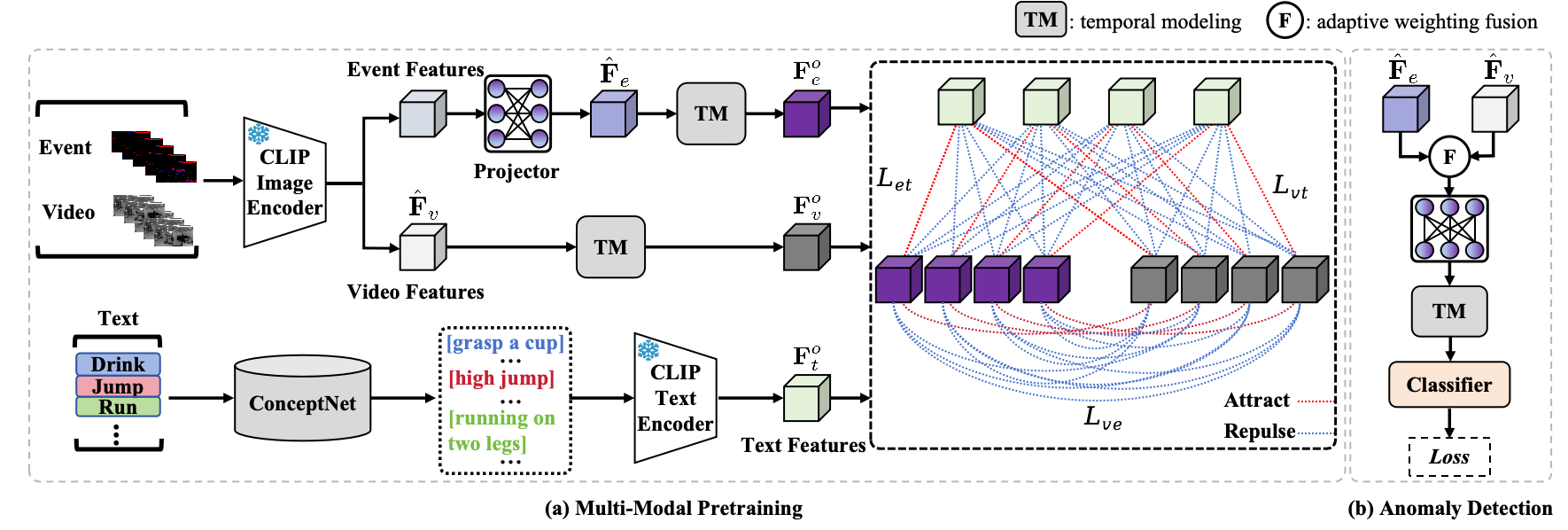}
    \caption{An overview of the proposed E-VAD framework via collaboration of visible and event frames, which comprises two key modules: (a) a multi-modal pretraining module and (b) an anomaly detection module. For event and video frames, features are initially extracted using the CLIP image encoder. Event features $\textbf{F}_e$ are projected to align with the latent space of CLIP image encoder. The projected event features $\hat{\textbf{F}}_e$ and video features $\hat{\textbf{F}}_v$ are then fed into a temporal modeling (TM) module to capture temporal dependencies and contextual interactions, yielding $\textbf{F}_e^o$ and $\textbf{F}_v^o$. For text processing, ConceptNet extracts semantically relevant concepts, encoded via the CLIP text encoder to obtain $\textbf{F}_t^o$. After acquiring these tri-modal features, pairwise contrastive losses ($L_{et}$, $L_{vt}$, $L_{ve}$) are computed to optimize the pretraining module. Post-pretraining, refined event features $\hat{\textbf{F}}_e$ are derived, followed by adaptive weighting fusion of event features $\hat{\textbf{F}}_e$ and video features $\hat{\textbf{F}}_v$. A projector adjusts these fused features for anomaly detection, after which the TM module captures residual temporal dependencies. Finally, anomaly scores are generated via a classifier, and the detection loss is computed for optimization.}
    \label{fig:overall_framework}
\end{figure*}

\subsection{Overview}

Let the visible video be $\mathcal{\mathbf{V}} = \{\mathbf{V}_1, \mathbf{V}_2,..., \mathbf{V}_T\}$ with T visible frames and the corresponding event stream contain asynchronous points $\{\mathbf{e}_n\}_{n=1}^N$ with $N$ points. The task is to estimate snippet-level anomaly scores under weak supervision, where only video-level labels $y\in \{0,1\}$ are provided. Raw events are converted into frame-like representations $\mathcal{E} = \{\mathbf{E}_i\}_{i=1}^T$ to enable efficient alignment with video and text encoders. Since VAD relies primarily on inter-frame temporal evolution rather than fine intra-frame timing, these representations remain sufficiently informative. Video-level textual categories are manually annotated and expanded via ConceptNet \cite{speer2013conceptnet}, providing language-based supervisory signals that anchor semantic concepts to event-induced motion patterns.

The proposed framework operates in two stages. A multi-modal pretraining module first aligns event, video, and textual embeddings to reduce cross-modal domain gaps. An anomaly detection network then adaptively fuses video and event features and models temporal dynamics to derive snippet-wise anomaly likelihood \cite{fan2024weakly}. In this design, events contribute temporally precise motion cues, videos provide spatial and contextual structure, and language supervision strengthens semantic discrimination between normal and abnormal behaviors. The complete architecture is shown in Figure \ref{fig:overall_framework}.

\subsection{Multi-Modal Pretraining for Event Representation Learning}

The core purpose of multi-modality pretraining is to obtain event features that are directly compatible with video features and textual guidance. Pretraining is composed of three components. The first component is an input encoding network which extracts representations from video, event, and text streams. The second component is a temporal modeling module that captures long-range and local dynamics. The third component is a contrastive training strategy which aligns the three modalities into a shared semantic space. This stage enables event features to carry not only motion cues but also anomaly relevant concepts interpretable by video and text signals.

\subsubsection{Input Encoding Network}

The input encoding network contains three paths for video, event, and text respectively. Each event element $\mathbf{e}_n$ is represented as $\{a,c,h,r\}$ where $(a,c)$ are spatial coordinates, $h$ is the timestamp, and $r$ is the polarity. Event streams are accumulated to produce event frames $\mathbf{E}_i \in \mathbb{R}^{C\times H \times W}$ in order to leverage the encoding strength of established image models. Video frames are fed into the same encoder to ensure spatial alignment across modalities.

The encoder produces video features $\hat{\mathbf{F}}_v \in \mathbb{{R}}^{L \times D}$ and event features $\mathbf{F}_e \in \mathbb{{R}}^{L \times D}$, where $L$ is the sequence length and $D$ is the feature dimension. Since the encoder is trained on visible data rather than event patterns, a domain shift exists. Therefore, a learnable projector is used to map event features into the latent space of visible embeddings:
 \begin{equation}
\hat{\mathbf{F}}_{e} = \text{Projector}(\mathbf{F}_e).
\end{equation}

To incorporate semantic knowledge, textual concepts for each anomaly category are first expanded by ConceptNet \cite{speer2013conceptnet}. The resulting concept list is encoded to obtain textual features $\mathbf{{F}}^{o}_{t}$. This process enhances discriminability among anomaly sub-classes by injecting structured prior knowledge.

\subsubsection{Temporal Modeling}
To capture temporal dependencies across modalities, a temporal modeling network is employed \cite{yan2024referred,pu2024learning,hu2024enhancing}. Linear projections generate $\mathbf{Q}, \mathbf{K}, \mathbf{V} \in \mathbb{{R}}^{m\times d_{k}}$  for both video and event sequences. Global temporal structure is modeled using standard scaled dot product attention:
\begin{equation}
    \mathbf{X}^g = \text{softmax}\left(\frac{ \mathbf{QK^\top}}{\sqrt{d_k}}\right)\mathbf{V}.
\end{equation}
Local variations and short term dynamics are captured by attention masking:
\begin{equation}
    \mathbf{X}^l = \text{softmax}\left(\frac{\mathbf{M} \times ( \mathbf{QK^\top})}{\sqrt{d_k}}\right)\mathbf{V},
\end{equation}
where mask entries $\mathbf{M}_{i,j} = 1$ when $(i,j)$ lies within a predefined temporal interval, otherwise $\mathbf{M}_{i,j}=-\infty$. Final modality-specific features $\textbf{F}_e^o$  and $\textbf{F}_v^o$ combine both temporal ranges using a learnable interpolation parameter $\mu$,
\begin{equation}
    \mathbf{F}^o = \mu \times \mathbf{X}^g + (1-\mu) \times \mathbf{X}^l.
\end{equation}
This gating mechanism enables capturing abrupt anomalies while maintaining long-term regularities.

\subsubsection{Contrastive Multi-Modality Training Strategy}

The pretraining stage aligns event features, video features, and text features in a unified embedding space. Cross-modal features of the same class are pulled closer and those of different classes are pushed apart using three contrastive objectives. The loss between modality $\rho$ and modality $q$ is defined as:
\begin{equation}
L_{\rho q} = -\frac{1}{N} \sum_{i=1}^{N} \log \frac{\exp(\text{sim}(\textbf{F}_{\rho}^{o(i)}, \textbf{F}_{q}^{o(i)}) / \tau)}{\sum_{j=1}^{N} \exp(\text{sim}(\textbf{F}_{\rho}^{o(i)}, \textbf{F}_{q}^{o(j)}) / \tau)},
\end{equation}
Where $\rho, q \in \{e, t, v\}$ and $\rho! = q$, sim(·, ·) denotes cosine similarity, and $\tau$ is the temperature parameter. The total pretraining objective is:
\begin{equation}
L_{total_{pre}} = \theta L_{et} + \beta L_{ve} + \gamma L_{vt}.
\end{equation}
Here, $\theta$, $\beta$, and $\gamma$ denote the respective weighting coefficients. This contrastive objective teaches event features to exhibit semantic consistency with both visible and textual features, enabling a more discriminative foundation for anomaly reasoning. Please notice that CLIP is not used as a pretrained event encoder. Instead, CLIP serves as a semantic anchor that provides a stable multi-modal embedding space. The contrastive training learns an event encoder that projects event representations into this space, effectively mitigating domain gap.

\subsection{Adaptive Weighting Fusion of Event and Video Modalities}
\begin{figure}
    \centering
    \includegraphics[width=0.6\linewidth]{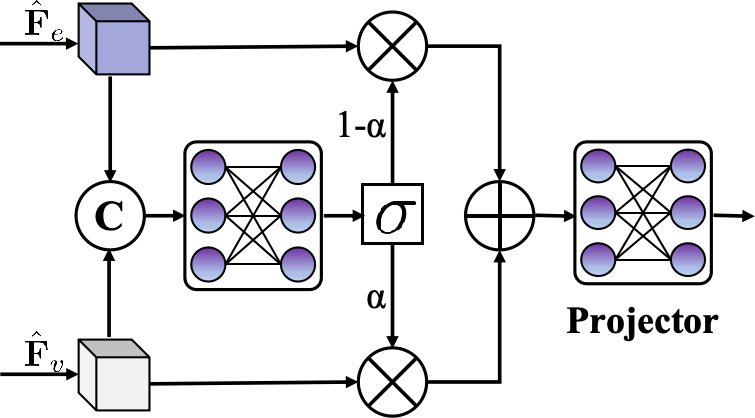}
    \caption{Architecture of the proposed adaptive weighting fusion module, which dynamically balances video features $\hat{\textbf{F}}_v$ and event features $\hat{\textbf{F}}_e$ through a learnable weighting mechanism.}
    \label{fig:fusion}
\end{figure}

Once semantic alignment is achieved, event and video features are combined within the detection framework. Instead of concatenation, fusion is conducted through a learnable weighting mechanism that assesses the relative reliability of each modality, as illustrated in Fig. \ref{fig:fusion}. The module computes a gating parameter $\alpha$ as
\begin{equation}
\alpha = \sigma(\textbf{W}_{\alpha} [\hat{\mathbf{F}}_v; \hat{\mathbf{F}}_e] + \textbf{b}_{\alpha}),
\end{equation}
where $\sigma(\cdot)$ denotes the sigmoid function, and $[\cdot;\cdot]$ represents feature concatenation. $\textbf{W}_{\alpha}$ and $\textbf{b}_{\alpha}$ are the weight parameters and bias of a linear layer, respectively. The fused representation is obtained by an element-wise weighting combination of both modalities:
\begin{equation}
\mathbf{F}_{f} = \alpha \odot \hat{\mathbf{F}}_v + (1 - \alpha) \odot \hat{\mathbf{F}}_e,
\end{equation}
followed by a linear projection to the desired output dimension:
\begin{equation}
\mathbf{F}_{out} = \textbf{W}_f \mathbf{F}_{f} + \textbf{b}_f.
\end{equation}
where $\textbf{W}_{f}$ and $\textbf{b}_{f}$ are the weight parameters and bias of the projection layer, respectively. The lightweight linear gating of AWF avoids the overfitting risk of high-capacity fusion modules, which is crucial under the low-sample, long-tail distributions typical of VAD. This fusion architecture ensures responsiveness to dynamic cues when events are reliable and dependence on visible cues when spatial detail is dominant. The output representation serves as input to anomaly prediction.

\subsection{Classifier for Snippet-level Anomaly Scoring}
To obtain more reliable results, both current and historical observations are considered to predict the anomaly scores by using snippet-level features $\{ \mathbf{F}_{out}^{s_i} \}_{i=1}^U\in \mathbf{F}_{out}$, where $U$ is the number of snippets. A causal convolution layer is adopted to achieve this goal, and it can be formulated as:
\begin{equation}
s_i=\sigma(f_t(\mathbf{{F}}_{out}^{s_i})),
\end{equation}
where $f_t$ is the causal convolution layer, $\sigma(\cdot)$ is the sigmoid function, and $s_i$ is the anomaly score of the $i^{th}$ snippet. Scores measure abnormality strength of each snippet and are used in weak supervision training.
\subsection{Loss Function}
Following work \cite{wu2021learning, pu2024learning, sultani2018real}, the final training objective integrates a multiple instance learning objective with a Kullback-Leibler divergence loss. The video-level prediction $\hat{y}_z$ is the mean of top-k snippet scores. For anomalous videos $k=\lfloor T/16+1\rfloor$. For normal videos $k=1$. The classification objective is:
\begin{equation}
    \mathcal{L}_{cs} = -\sum_{z=1}^{Z} {y}_zlog(\hat{y}_z),
\end{equation}
where $y_z$ is the video-level ground-truth label. $Z$ is the number of training samples. Additionally, semantic consistency is strengthened using Kullback-Leibler divergence:
\begin{equation}
    \mathcal{L}_{kd} = \mathbb{E}_{p \sim p(v)} \left[ \log p^{v2t}(v) - \log q^{v2t}(v) \right],
\end{equation}
here $p^{v2t}{(v)}$ denotes similarity scores between vision features and textural features, and $q^{v2t}{(v)}$ represents semantic consistency labels (1 for positive pairs, 0 otherwise). The final optimization objective combines losses with weighted summation:
\begin{equation}
\mathcal{L} = \mathcal{L}_{ce} + \lambda \mathcal{L}_{kd},
\end{equation}
where $\lambda$ balances the terms. This formulation encourages the network to discover the minimal abnormal regions that explain the label, while benefiting from event-driven temporal dynamics and textual semantic regularization.

\begin{table*}[]
    \centering
    \setlength{\tabcolsep}{2pt}
    \caption{Detailed comparison of existing public video anomaly detection datasets and the proposed TJUTCM Pha dataset. The table highlights differences in year, project link, modality (video/event), task, the number of events, the number of video frames, and notes, showing that TJUTCM Pha provides the first real-world visible–event benchmark for anomaly detection task.}
    \begin{tabular}{ccccccccc} 
    \toprule
   {\textbf{Dataset}} & {\textbf{Year}} & {\textbf{Project}} & {\textbf{Modality}} &\textbf{Task} & {\textbf{\#Events }} & {\textbf{\#Frame}} & \textbf{Notes for Events} \\
    \hline 
 \textbf{Subway Entrance} \cite{adam2008robust} & 2008 & \href{https://vision.eecs.yorku.ca/research/anomalous-behaviour-data/}{\textcolor{red}{URL}} &Video& VAD &0&137k & -\\ 
    \textbf{Subway Exit} \cite{adam2008robust} & 2008 & \href{https://vision.eecs.yorku.ca/research/anomalous-behaviour-data/}{\textcolor{red}{URL}}  & Video& VAD&0& 72k   &  -\\
    \textbf{CUHK Avenue} \cite{lu2013abnormal} & 2013 & \href{https://www.cse.cuhk.edu.hk/leojia/projects/detectabnormal/dataset.html}{\textcolor{red}{URL}} & Video & VAD & 0& 31k   &-\\ 
    \textbf{UCSD Ped2} \cite{mahadevan2010anomaly} & 2013 & \href{http://www.svcl.ucsd.edu/projects/anomaly/dataset.htm}{\textcolor{red}{URL}}  & Video & VAD & 0& 5k  &- \\
    \textbf{UCSD Ped1} \cite{mahadevan2010anomaly} & 2013 & \href{http://www.svcl.ucsd.edu/projects/anomaly/dataset.htm}{\textcolor{red}{URL}}  & Video & VAD &0& 14k   &-\\ 
    \textbf{XD} \cite{wu2020not} & 2020 &\href{https://roc-ng.github.io/XD-Violence/}{\textcolor{red}{URL}}  & Video & VAD & 0 & 93,405k   & \\ 
    \hline
     \textbf{DVS-Gesture} \cite{amir2017low} & 2017 & 
\href{http://research.ibm.com/dvsgesture/}{\textcolor{red}{URL}}& Event &Gesture Recognition & 0.5B & 0 &  Real\\
   \textbf{ASL-DVS} \cite{bi2019graph} & 2020 &\href{https://github.com/PIX2NVS/NVS2Graph}{\textcolor{red}{URL}}& Event &Classification & 2B & 0 & Real & \\ 
    \textbf{MEVDT} \cite{dong2023bullying10k} & 2023 &\href{https://doi.org/10.7302/d5k3-9150}{\textcolor{red}{URL}} & Image + Event &Object Detection and Tracking & 5.5M &1.3k & Real \\ 
    \hline
    \textbf{UCF-Crime} \cite{sultani2018real, qian2025ucf} & 2025 & \href{https://github.com/YBQian-Roy/UCF-Crime-DVS}{\textcolor{red}{URL}}  & Video + Event& VAD & 101.4B & 13,824k &\makecell{Re-recorded event streams inheriting \\ limitations of visible video sensing}\\
    \textbf{\makecell{ShanghaiTech Campus-Event \\ (Ours)}} \cite{liu2018future} & 2025 & \href{https://svip-lab.github.io/dataset/campus_dataset.html}{\textcolor{red}{URL}}  & Video + Event & VAD & 5.1B & 317k  & \makecell{Simulation-based event \\ extension}\\
    \textbf{TJUTCM Pha (Ours)} & 2025 & URL\tablefootnote{The project link will be release later.}  & Video + Event & VAD & 6.3B &377k &\makecell{\textbf{First real-world} visible–event \\ dataset designed for VAD}\\
    \bottomrule
    \end{tabular}
    \label{tab:datasets}
\end{table*}

\section{TJUTCM Pha Dataset}\label{sec:data}

Recent progress in VAD has been largely driven by improvements in model architectures built upon visible-light video benchmarks. However, the sensing modality itself has remained almost unchanged for more than a decade. As summarized in Table~\ref{tab:datasets}, most widely used VAD datasets, including Subway Entrance/Exit~\cite{adam2008robust}, CUHK Avenue~\cite{lu2013abnormal}, UCSD Ped1/Ped2~\cite{mahadevan2010anomaly}, XD~\cite{sultani2018real}, and ShanghaiTech Campus~\cite{liu2018future}, rely exclusively on frame-based videos and therefore inherit the intrinsic limitations of conventional cameras, such as motion blur, illumination sensitivity, and temporal discretization.

The recently introduced UCF-Crime-DVS~\cite{qian2025ucf} represents an important step toward incorporating event information into anomaly detection by extending the large-scale UCF-Crime benchmark with event streams. Its primary contribution lies in enabling scalable multi-modal evaluation on long-duration videos. However, the event data in UCF-Crime-DVS are generated by re-recording videos displayed on a monitor, which means that the resulting events remain perceptually dependent on the original video content. As a consequence, the event streams inherit the temporal and photometric constraints of visible sensing and do not fully reflect the asynchronous dynamics, high dynamic range, or noise characteristics of real event cameras.

To complement existing resources and address this fundamental gap, TJUTCM Pha is introduced as the first real-world benchmark that synchronously captures visible video frames and event streams using a single hybrid event camera for the VAD task. Unlike replay-based or simulated datasets, TJUTCM Pha is collected directly from operational pharmaceutical laboratories and production lines, preserving authentic event generation mechanisms driven by real motion and illumination changes. This design enables the study of anomaly detection beyond the perceptual ceiling imposed by video-based sensing and provides a realistic testbed for evaluating event-enhanced VAD models under various challenging conditions.

\subsection{Data Acquisition Protocol and Annotation Principles}

All data are recorded using a DAVIS346 hybrid event camera deployed in real pharmaceutical production and laboratory environments. The sensor simultaneously outputs synchronized visible frames and asynchronous event streams, ensuring precise temporal correspondence between the two modalities while retaining the native sensing characteristics of each. Each sequence lasts approximately 30 seconds and captures uninterrupted operational processes, allowing abnormal behaviors to emerge naturally at arbitrary temporal locations.

\begin{figure}
    \centering
    \includegraphics[width=0.95\linewidth]{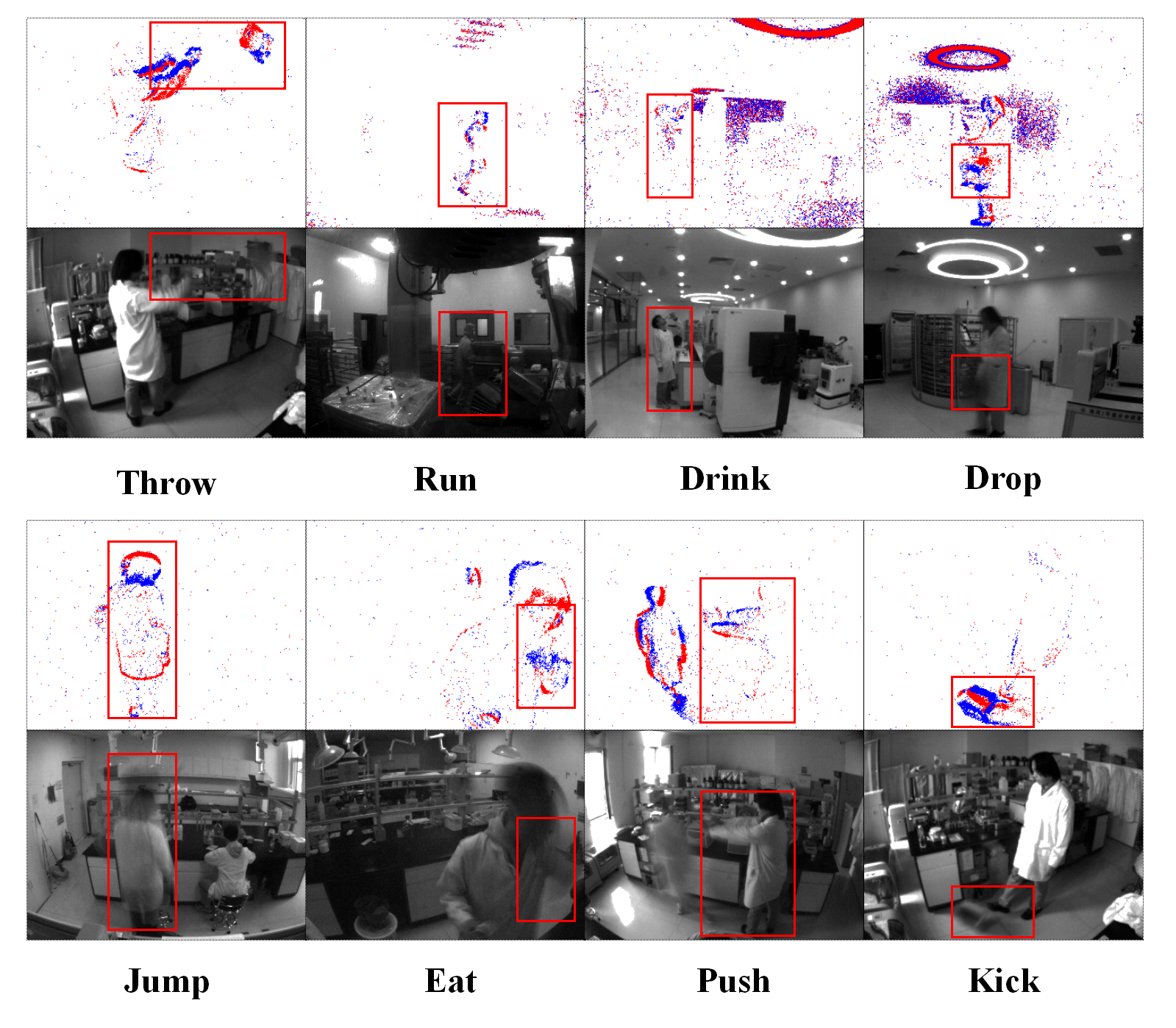}
    \caption{Representative synchronized visible and event frames from the proposed TJUTCM Pha dataset samples. The abnormal categories encompass throw, run, drink, drop, jump, eat, push, kick, etc.}
    \label{fig:repre}
\end{figure}

Annotations follow a two-level semantic scheme. Normal samples correspond to compliant laboratory operations, routine material handling, and standard pharmaceutical procedures. Abnormal samples include actions that violate safety regulations or operational protocols, such as throwing, pushing, kicking, running, jumping, drinking, eating, object dropping, and other hazardous behaviors. These anomalies are intentionally diverse and subtle, reflecting realistic safety risks encountered in pharmaceutical environments. Representative samples from both modalities are shown in Fig.~\ref{fig:repre}.

\subsection{Dataset Characteristics and Statistical Properties}

TJUTCM Pha is designed to support robust evaluation of multi-modal anomaly detection under realistic sensing conditions. Its key characteristics are summarized below:

\begin{enumerate}
    \item \textbf{Real-world visible–event sensing}: Both video frames and event streams are captured directly from physical scenes, ensuring genuine asynchronous dynamics and eliminating perceptual dependence on replayed videos.
    
    \item \textbf{Industrial scene diversity}: Data are collected across ten heterogeneous pharmaceutical areas with varying illumination, different motion speeds, occlusion levels, and equipment layouts, encouraging generalization across complex operational settings.
    
    \item \textbf{Continuous temporal evolution}: Each sequence contains uninterrupted behavior trajectories, with anomalies occurring sparsely and unpredictably over time. This structure emphasizes temporal reasoning rather than isolated frame discrimination.
    
    \item \textbf{Natural event activation}: Event density emerges organically from motion and lighting variations, resulting in non-uniform temporal activity patterns that reflect true event camera behavior and pose challenges for temporal modeling.
    
    \item \textbf{Strict cross-modal synchronization}: Visible frames and event streams are temporally aligned to support contrastive learning, cross-modal fusion, and fine-grained temporal correspondence in anomaly detection.
\end{enumerate}

In summary, TJUTCM Pha complements existing datasets by introducing authentic visible–event sensing into the VAD domain. While UCF-Crime-DVS enables large-scale benchmarking with replay-based events, TJUTCM Pha provides a fundamentally different and indispensable resource, i.e., a real-world, industrial-scale dataset that captures genuine event dynamics alongside visible videos. This dataset enables the community to systematically investigate sensing-aware anomaly detection and lays a solid foundation for developing robust multi-modal VAD systems beyond the limitations of video-based perception.


\section{Experiments}\label{sec:expe}

\subsection{Datasets and Evaluation Metrics}

The effectiveness and generalizability of E-VAD are assessed using two widely used anomaly detection benchmarks and one newly established pharmaceutical behavior anomaly dataset. These datasets cover a broad spectrum of scenarios ranging from public campus scenes to complex industrial environments.

\begin{enumerate}
    \item \textbf{UCF-Crime Dataset}
The UCF-Crime dataset contains 1,900 untrimmed surveillance videos, spanning 13 categories of real-world anomalous events such as robbery, explosion, and accidents. Following the standard weak supervision protocol, 1,610 videos are used for training and 290 for testing, where only video-level labels are employed during training and frame-level annotations serve exclusively for evaluation. Event streams are obtained via the UCF-Crime-DVS extension \cite{qian2025ucf}. However, UCF-Crime-DVS generates events by re-recording videos displayed on a monitor, causing the event data to inherit the degradations and frame-rate constraints of the original visible footage rather than the sensing characteristics of real event cameras. Consequently, it lacks genuine microsecond temporal dynamics, high dynamic range, and naturally sparse motion-triggered responses, making it substantially less realistic than authentic event datasets for multi-modal anomaly detection.

    \item \textbf{ShanghaiTech Campus Dataset}
The ShanghaiTech Campus dataset consists of 437 surveillance videos collected across 13 campus scenes, with 238 videos used for training and 199 for testing. 
To enable event–video multi-modal evaluation on this widely adopted benchmark, an event-based extension, termed \textbf{ShanghaiTech-Event}, is constructed in this work by simulating event streams from the original videos using the public conversion framework in \cite{lin2022dvs}. 
This extension provides a standardized and reproducible multi-modal variant of ShanghaiTech, facilitating fair comparison between video-only and event-augmented VAD methods. Noted that, as the event streams are synthesized from clean visible-light footage rather than captured by physical event sensors, ShanghaiTech-Event does not exhibit the full spectrum of challenges inherent to real-world event acquisition, such as illumination-induced noise, motion-triggered sparsity, or sensor-level artifacts. Consequently, while ShanghaiTech-Event serves as a useful controlled benchmark for studying cross-modal modeling behavior, it lacks the complex sensing conditions that characterize genuine event-based data captured in real environments. Despite these limitations, ShanghaiTech-Event remains valuable for isolating modeling effects under controlled sensing conditions, making it suitable for ablation and cross-modal mechanism analysis.

    \item \textbf{TJUTCM Pha Dataset}
We construct TJUTCM Pha to evaluate anomaly detection in pharmaceutical laboratory and production environments. Eighty-five percent of the samples are used for training and the remaining fifteen percent for testing. The dataset exhibits fine-grained manipulation actions, occlusions, lighting variation, and subtle anomalies. This enables rigorous assessment of multi-modal temporal discrimination in real scenes that require precise anomaly understanding.
\end{enumerate}

\textbf{Evaluation Metrics.} Two complementary metrics are adopted. AUC evaluates frame-level anomaly localization capability, where higher values indicate better separability between normal and abnormal segments. FAR measures false alarms at a fixed threshold of 0.5, reflecting system stability and practical reliability. Throughout all tables, the best results are shown in bold, the second-best results are underlined, and the third-best results are highlighted in blue.

\subsection{Implementation Details}

Event, video and text features are extracted using CLIP ViT-L/14, and projected into a shared embedding space of 768 dimensions to ensure cross-modal semantic alignment. The temporal reasoning module employs a latent dimension of 128 and uses a local interval length of five frames for contextual aggregation. The fusion coefficient is initialized to 0.5 to ensure balanced multi-modal contribution at early training stages. The maximum processed sequence during training is limited to 200 frames to enable uniform snippet sampling for videos of varying length.

Model optimization uses the Adam optimizer with an initial learning rate of $1 \times 10^{-3}$ and a batch size of 512 for 1000 epochs. Gradient clipping is applied with an upper bound of 1.0 to ensure stable backpropagation. These configurations are selected to balance training efficiency, representation compactness and generalization robustness under multi-modal weak supervision.

\subsection{Comparison with State-of-the-Art Methods}
Quantitative comparisons on ShanghaiTech, UCF-Crime and TJUTCM Pha are reported in Tables \ref{tab:ShanghaiTech}, \ref{tab:UCF} and \ref{tab:TJUTCM Pha}, respectively. Competing methods include both video-only and multi-modal approaches. Methods annotated with “*” are re-trained using identical ``video + event" configurations to ensure strict fairness; therefore, all reported gains are attributed to not only data differences but also algorithmic superiority.
\begin{table}[ht!]
\centering
\caption{Performance comparison between the proposed E-VAD and SOTA methods on the ShanghaiTech Campus dataset under weakly-supervised settings. ``\textbf{Bold}" numbers denotes the best experimental performance. ``*" denotes the method that we re-train its code utilizing both the video and event modality in the experiments.}
\label{tab:my_table}
\begin{tabular}{ccccc}
\toprule
 \textbf{Method} & \textbf{Feature} & \textbf{Modality} & \textbf{AUC}(\%) & \textbf{FAR}(\%) \\
\midrule
{MIL-Rank} \cite{sultani2018real}& C3D& video & 86.30 & 0.15 \\
 {IBL} \cite{zhang2019temporal}& C3D &video & 82.50 & 0.10 \\
 {GCN} \cite{zhong2019graph}& TSN &video & 84.44 & - \\
  CLAWS \cite{zaheer2020claws}& C3D &video & 89.67 & - \\
  MIST \cite{feng2021mist}& I3D &video & 94.83 & 0.05 \\
  CRFD \cite{wu2021learning}& I3D &video & 97.48 & - \\
  RTFM \cite{tian2021weakly}& I3D &video & 97.21 & - \\
MSL \cite{li2022self}& VideoSwin &- & 97.32 & - \\
  NL-MIL \cite{park2023normality} & I3D &video & 97.43 & - \\
  S3R\cite{wu2022self} & I3D &video & 97.48 & - \\
  Cho et al.\cite{cho2023look} & I3D &video & 97.60 & - \\
  AR-Net\cite{wan2020weakly} & &video+flow & 91.24 & 0.10 \\
  UML\cite{lv2023unbiased} & X-CLIP &video & 96.78 & - \\
  
   ProDisc \cite{zhu2025prodisc}& I3D &video & 97.98 & - \\
    
 PEL4VAD\cite{pu2024learning} & I3D &video & 98.14 & \textbf{0.00} \\ \hline
   UML\cite{lv2023unbiased}* & X-CLIP &video+event & 96.63 & - \\
   ProDisc \cite{zhu2025prodisc}*& I3D &video+event & 98.02 & - \\
   PEL4VAD\cite{pu2024learning}* & I3D &video+event & 98.21 & 0.13 \\  \hline
  \textbf{E-VAD(Ours)} & CLIP &video+event & \textbf{98.67} & \textbf{0.00} \\
\bottomrule
\end{tabular}
\label{tab:ShanghaiTech}
\end{table}

\begin{itemize}
\item \textbf{Results on ShanghaiTech dataset.}
Performance comparisons on the ShanghaiTech dataset are presented in Table \ref{tab:ShanghaiTech}. The results indicate that E-VAD achieves the highest performance among all competing methods, reaching 98.67\% AUC and 0.00\% FAR. For example, relative to strong baselines such as ProDisc-VAD and PEL4VAD, E-VAD delivers AUC improvements of 0.65\% and 0.46\%, respectively. These improvements demonstrate that jointly exploiting event streams and visible videos substantially enhances anomaly detection, as the temporally precise motion cues captured by event data effectively complement the spatial structures preserved in visible frames. Although several prior methods obtain minor performance gains after incorporating event information, their overall results remain consistently lower than those of E-VAD. As an illustration, PEL4VAD* achieves an AUC of 98.21\%, still 0.46\% below E-VAD. This performance gap highlights that the proposed multi-modal pretraining module yields more discriminative event representations, and that the adaptive weighting fusion strategy integrates motion-salient signals more effectively, particularly under adverse conditions such as illumination fluctuations and background clutter.
\item \textbf{Results on UCF-Crime dataset.}
Table \ref{tab:UCF} presents the experimental results on the UCF-Crime dataset, where E-VAD is evaluated against prior works including VADCLIP, PEL4VAD, and their corresponding multi-modal variants. The results show that E-VAD achieves an AUC of 88.75\%, which is comparable to the 89.04\% reported by ITC while surpassing all remaining methods. Although its FAR matches that of GCN and MIST, the AUC of E-VAD exceeds these approaches by 6.63\% and 6.45\%, respectively. An interesting trend is that integrating event streams into VADCLIP, PEL4VAD, and UML leads to a slight decline in their AUC values, primarily because screen-projected events not only inherit the intrinsic weaknesses of the visible modality but also introduce extra display and recording noise during the filming process, resulting in a degraded SNR and compromised anomaly discriminability. Despite this, E-VAD consistently outperforms all alternatives in both AUC and FAR, underscoring the inherent difficulty faced by video-centric approaches when detecting subtle, short-lived abnormalities in cluttered environments. Benefiting from the higher temporal resolution and wider dynamic range of event data, the proposed method effectively suppresses static background clutter and reduces noise, while the pretraining framework provides strong cross-modal semantic priors that further enhance detection accuracy.
\item \textbf{Results on TJUTCM Pha dataset.} 
The generalization capability of E-VAD is rigorously evaluated on the fine-grained TJUTCM Pha dataset under multi-modal settings. As shown in Table \ref{tab:TJUTCM Pha}, E-VAD attains 82.25\% AUC with a 0.00\% FAR, outperforming existing multi-modal approaches by substantial margins of 8.47\%–10.11\%. In contrast to methods such as ProDisc-VAD*, VADCLIP*, and PEL4VAD*, which predominantly depend on video appearance features supplemented by event streams, E-VAD benefits from a dedicated cross-modal pretraining strategy aligned with video–text pairs and an adaptive weighting fusion module. These two components jointly enhance the ability of the model to extract and utilize dynamic-sensitive event cues, enabling precise characterization of subtle manipulations and fine-grained operational behaviors typical of laboratory environments. Consequently, E-VAD demonstrates markedly stronger robustness in identifying anomalous patterns across diverse operational conditions, highlighting its effectiveness in scenarios where traditional video-centric methods struggle to preserve temporal fidelity and capture nuanced motion signatures.
\end{itemize}

Overall, evaluations across three distinct datasets consistently show that E-VAD achieves robust performance in both coarse-grained public surveillance and fine-grained industrial anomaly detection scenarios.

\begin{table}[h!]
\centering
\caption{Experimental comparison of the proposed E-VAD method with existing SOTA methods on the UCF-Crime dataset.}
\label{tab:UCF}
\begin{tabular}{ccccc}
\toprule
 \textbf{Method} & \textbf{Feature} & \textbf{Modality}& \textbf{AUC}(\%) & \textbf{FAR}(\%) \\
\midrule
MIL-Rank\cite{sultani2018real}  & C3D &video & 75.41 & 1.9 \\
  IBL\cite{zhang2019temporal}  & C3D &video & 78.66 & - \\
  Motion-Aware\cite{zhu2019motion}  & PWC &flow & 79.00 & - \\
  GCN\cite{zhong2019graph}  & TSN &video & 82.12 & \textbf{0.1} \\
  MIST\cite{feng2021mist}  & I3D &video & 82.30 & \underline{0.13} \\
  HL-Net\cite{wu2020not}  & I3D &video & 82.44 & - \\
  MS-BSAD\cite{zhen2021multi}  & I3D &video & 83.53 & - \\
  RTFM\cite{tian2021weakly}  & I3D &video & 84.30 & - \\
  CRFD\cite{wu2021learning}  & I3D &video & 84.89 & 0.72 \\
  DDL \cite{pu2022locality} & I3D &video & 85.12 & - \\
  MSL\cite{li2022self}  & I3D &video & 85.30 & - \\
  MLAD \cite{zhang2022weakly} & I3D &video & 85.47 & 7.47 \\
  NL-MIL\cite{park2023normality}  & I3D &video & 85.63 & - \\
  S3R\cite{wu2022self}  & I3D &video & 85.99 & - \\
  Cho et al. \cite{cho2023look} & I3D &video & 86.10 & - \\
  CUPL \cite{zhang2023exploiting} & I3D &video & 86.22 & - \\
  UML\cite{lv2023unbiased}  & X-CLIP &video & 86.75 & - \\
 PEL4VAD\cite{pu2024learning} & I3D &video & {86.76} & 0.43\\
 VADCLIP\cite{wu2023vadclip} & CLIP &video & \textcolor{blue}{88.02} & -\\ 
 MSF \cite{qian2025ucf} & SNNs & event & 65.10 & 3.27 \\
  ITC\cite{liu2024injecting} & CLIP & video &\textbf{89.04} & - \\ \hline
   VADCLIP\cite{wu2023vadclip}* & CLIP &video +event& 87.94 & -\\
    PEL4VAD\cite{pu2024learning}* & I3D &video+event & {87.91} & 1.134\\
       UML\cite{lv2023unbiased}*  & X-CLIP &video+event & 85.95 & - \\ \hline
  \textbf{E-VAD(Ours)} & CLIP & video+event & \underline{88.75} & \textcolor{blue}{0.40}\\
\bottomrule
\end{tabular}
\end{table}

\begin{table}[h!]
\centering
\caption{Comparison between the proposed E-VAD with representative SOTA algorithms on the proposed TJUTCM Pha dataset.}
\label{tab:TJUTCM Pha}
\begin{tabular}{ccc}
\toprule
 \textbf{Method}  & \textbf{AUC}(\%) & \textbf{FAR}(\%) \\
\midrule
   MGFN\cite{chen2023mgfn}* &  67.34 & - \\
  CLIP-TSA\cite{joo2023clip}* &  69.11 & - \\
   BN-WVAD\cite{zhou2024batchnorm}* &  71.85 & - \\
   PEL4VAD\cite{pu2024learning}* & 72.14 & \textbf{0.00} \\
    VADCLIP \cite{wu2023vadclip}* &  73.62 & - \\ 
  ProDisc \cite{zhu2025prodisc}* & 73.78 & - \\ \hline
  \textbf{E-VAD(Ours)} & \textbf{82.25} & \textbf{0.00} \\
\bottomrule
\end{tabular}
\end{table}

\subsection{Ablation Studies}

\begin{table}[htbp]
\centering
\caption{Ablation studies on the component analysis, including video, event, and MMP modules.}
\begin{tabular}{ccc|cc}
\toprule
\textbf{Video} & \textbf{Event} & \textbf{MMP} & \textbf{AUC}(\%) & \textbf{FAR}(\%) \\
\midrule
\checkmark & - & - & 75.30 & 0.21 \\
- & \checkmark & - & 78.15 & 0.14 \\
- & \checkmark & \checkmark & {79.79} & 0.00 \\
\checkmark & \checkmark & - & 80.91 & 0.04 \\
\checkmark & \checkmark & \checkmark & \textbf{82.25} & 0.00 \\
\bottomrule
\end{tabular}
\label{tab:ablation_modalities}
\end{table}

Ablation experiments on TJUTCM Pha analyze component contribution, fusion strategy, temporal modeling mechanisms, and event representations.
\subsubsection{Component Contribution}
Table \ref{tab:ablation_modalities} summarizes the contributions of the three key components, i.e., video, event, and MMP, to overall anomaly detection performance. The video baseline utilizes only visible frames for visual recognition, while the event baseline relies solely on event-driven inputs and employs the CLIP image encoder for feature extraction. In contrast, MMP introduces the proposed multi-modal pretraining strategy designed to enhance event representation quality. The results indicate that the video-only baseline achieves 75.30\% AUC, whereas the event-only approach reaches 78.15\%, demonstrating the inherent robustness of event data under challenging conditions such as cluttered backgrounds and extreme illumination. Importantly, event integrating the MMP module improves performance to 79.79\%, confirming its capability to extract more discriminative event features. Fusing videos and events further boosts the AUC to 80.91\%, highlighting the complementary nature of spatial information from videos and temporal dynamics from events. Finally, combining MMP with both visual modalities yields the highest AUC of 82.25\%, indicating that the proposed multi-modal pretraining effectively narrows cross-domain gaps among video, event, and text representations while enhancing the transferability of aligned semantic cues. This progression demonstrates the ability of the framework to fully exploit multi-modal information for superior weakly-supervised anomaly detection.

\subsubsection{Fusion Strategy Evaluation}

To assess the effectiveness of the proposed AWF mechanism, we conduct comparisons against several representative fusion strategies. \textbf{CAF} adopts a cross-attention formulation in which event features guide the selection of relevant video information. \textbf{CL} performs straightforward concatenation of video and event features followed by linear projection. \textbf{WF} introduces static and dynamic weighting via an attention-based MLP, whereas \textbf{MLPF} leverages a deeper multi-layer perceptron to model nonlinear cross-modal dependencies. \textbf{SWF} employs a single learnable scalar to control the relative importance of each modality. As reported in Table \ref{tab:fusion_ablation}, AWF achieves the best performance with an AUC of 82.25\%, surpassing CAF (64.95\%), CL (74.50\%), WF (76.38\%), MLPF (78.86\%), and SWF (80.92\%). These results demonstrate that AWF effectively adjusts modality contributions according to scene complexity, leading to more reliable and context-sensitive fusion compared with conventional static or attention-based approaches. Besides, the lightweight linear gating of AWF avoids the overfitting risk commonly observed in high-capacity fusion modules, a critical advantage under the low-sample, long-tail distribution characteristics of VAD.
\begin{table}[htbp]
\centering
\caption{Ablation study on different multi-modal feature fusion modules. }
\label{tab:fusion_ablation}
\begin{tabular}{l c c}
\toprule
\textbf{Method} & \textbf{AUC}(\%) & \textbf{FAR}(\%) \\
\midrule
CAF  &  64.95 & 0.34 \\
CL  & 74.50 & 0.22 \\
WF  & 76.38 & 0.08 \\
MLPF  & 78.86 & 0.05 \\
SWF  & 80.92 & 0.00 \\ \hline
\textbf{AWF(Ours) }&\textbf{ 82.25} & \textbf{0.00}\\

\bottomrule
\end{tabular}
\end{table}

\subsubsection{Temporal Modeling Mechanism}
As shown in Table \ref{tab:temporal}, we compare various temporal modeling mechanisms derived from other works, including various version of transformer, attention block, local and global mechanism, etc. The combination of cross-frame relations and transformer architecture achieves the best result with an AUC of 82.25\%. Compared to other strategies, this approach better captures both short-term and long-term temporal dependencies, effectively modeling the dynamic evolution of abnormal behaviors.
\begin{table}[htbp]
\centering
\caption{Evaluation of various temporal modeling mechanisms, including transformer-based, attention block, and hybrid approaches.}
\label{tab:temporal}
\begin{tabular}{l c c}
\toprule
\textbf{Temporal Mechanism} & \textbf{AUC}(\%) & \textbf{FAR}(\%) \\
\midrule
Dilated + Transformer~\cite{URDMU_zh}                          & 68.20 & 0.0012 \\
Dilated + Nonlocal~\cite{tian2021weakly}                       & 73.27 & 0.00   \\
Positional + Transformer~\cite{lv2023unbiased}                 & 72.93 & 0.00   \\
Nonlocal + Conv\cite{tu2023implicit}                          & 70.04 & 0.0002 \\
Attention + Transformer~\cite{tang2024predformer}              & 65.22 & 0.011  \\
Nonlocal + Transformer~\cite{yan2024referred}                  & 67.70 & 0.112  \\
Attention + Nonlocal~\cite{lu2023vdt} & 65.34 & 0.0028 \\
Gate + Transformer~\cite{hu2024enhancing}                          & 72.38 & 0.00   \\
Focal + Nonlocal~\cite{wasim2023video}                        & 69.59 & 0.00   \\
Guide + Transformer~\cite{yang2024fresco}                      & 65.30 &   0.0030 \\
Positional + Transformer~\cite{qian2024momentor}               & 68.97 & 0.0015 \\ \hline
CrossFrame + Nonlocal~\cite{pu2024learning}                  & \textbf{82.25} & \textbf{0.00}   \\
\bottomrule
\end{tabular}
\end{table}

\begin{table}[t]
\centering
\caption{Comparison of different event representations on the TJUTCM Pha dataset. }
\small
\begin{tabular}{lcc}
\hline
\textbf{Event Representation} & \textbf{AUC}(\%) & \textbf{FAR}(\%) \\
\hline

Voxel Grid \cite{gallego2020event}             & 66.40          & 1.00        \\
Time Surface \cite{gallego2020event}              & 71.87          & 0.15        \\
Event Frame (ours)     & \textbf{78.15} & \textbf{0.14} \\
\hline
\end{tabular}

\label{tab:event_representation}
\end{table}

\begin{figure*}
    \centering
\includegraphics[width=0.9\linewidth]{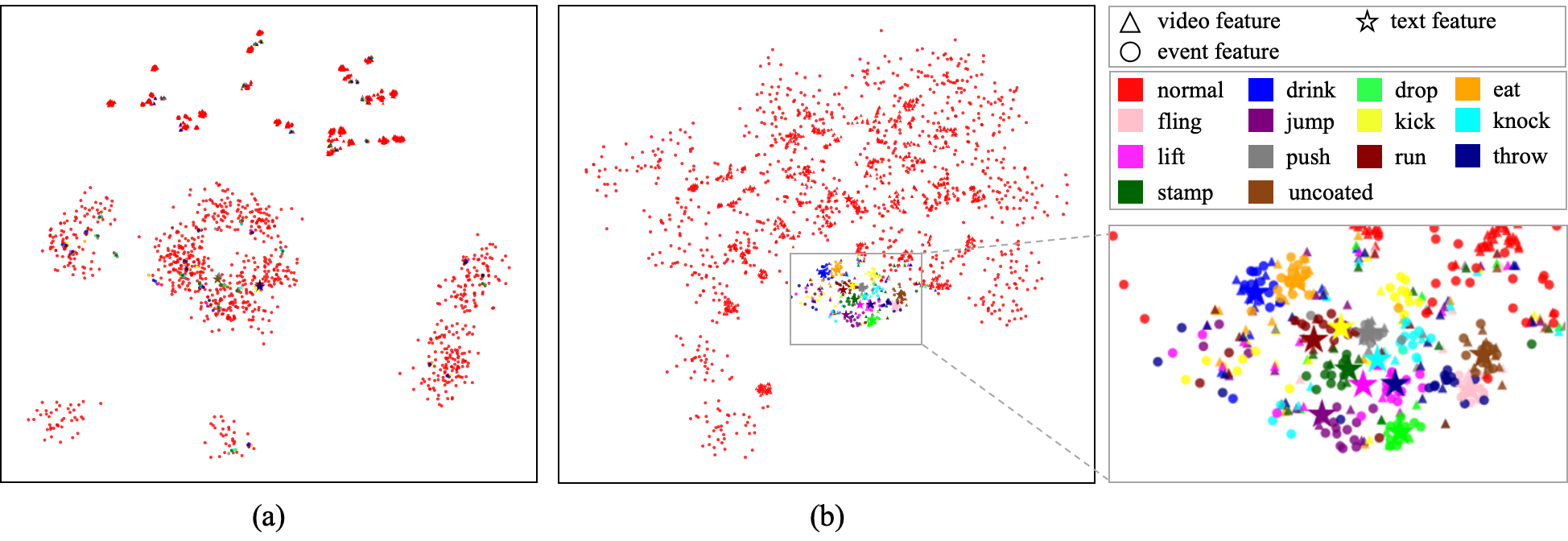}
    \caption{Visualization of feature embeddings before and after multi-modal pretraining via t-SNE\cite{maaten2008visualizing} on the proposed TJUTCM Pha dataset. Post-training results reveal enhanced alignment and tighter clustering of event, video, and text features within a shared semantic space.}
    \label{fig:tsne}
\end{figure*}

\begin{figure*}
    \centering
    \includegraphics[width=0.95\linewidth]{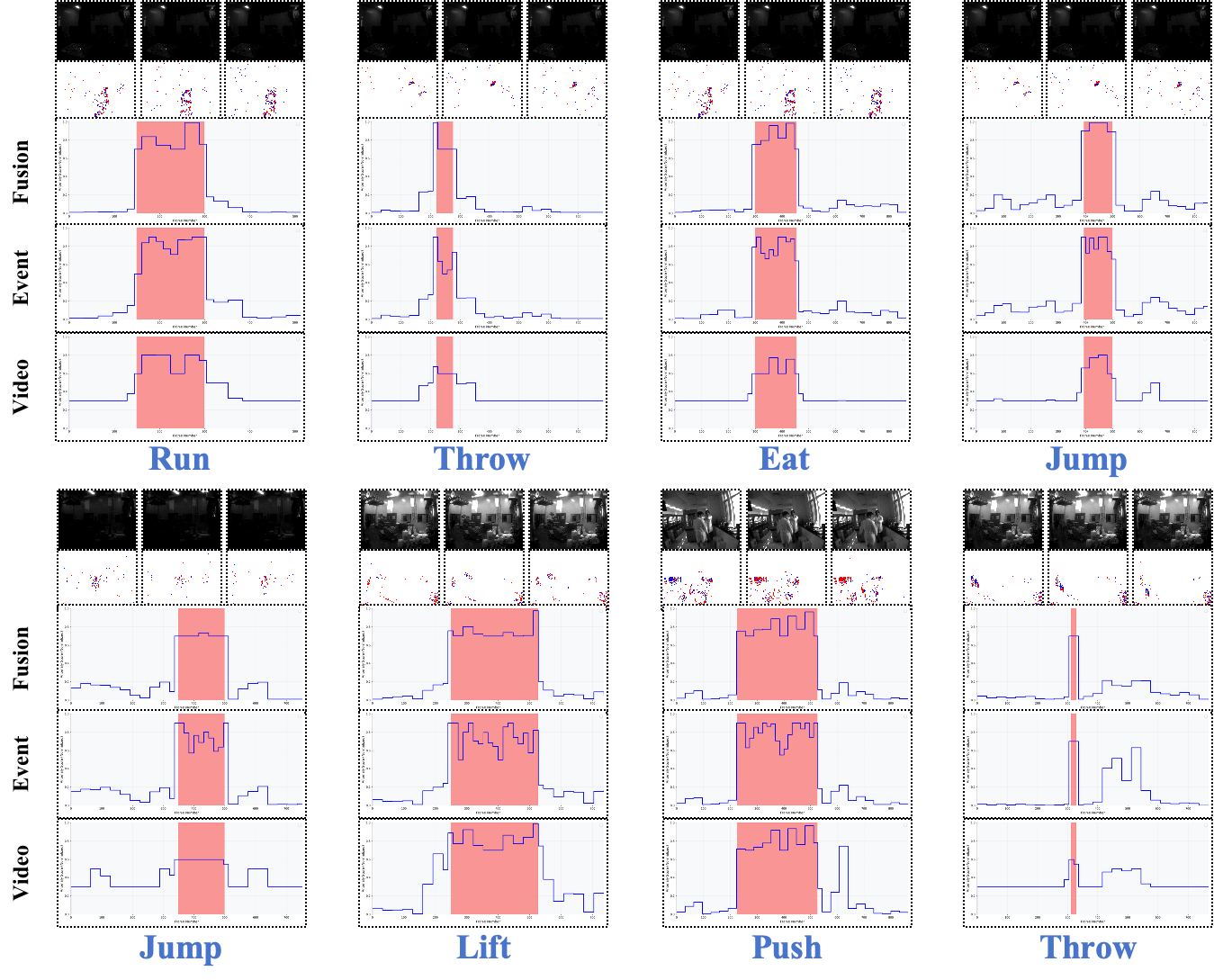}
    \caption{The frame-level visualizations for the abnormal detection of the proposed E-VAD method. Each example includes sampled video frames, event frames, and the corresponding anomaly score curves from the video branch, the event branch, and the fusion branch. The red shaded region denotes the annotated abnormal interval, providing a reference for the temporal alignment of the three score curves.}
    \label{fig:vis}
\end{figure*}

\begin{figure}
    \centering
    \includegraphics[width=0.6\linewidth]{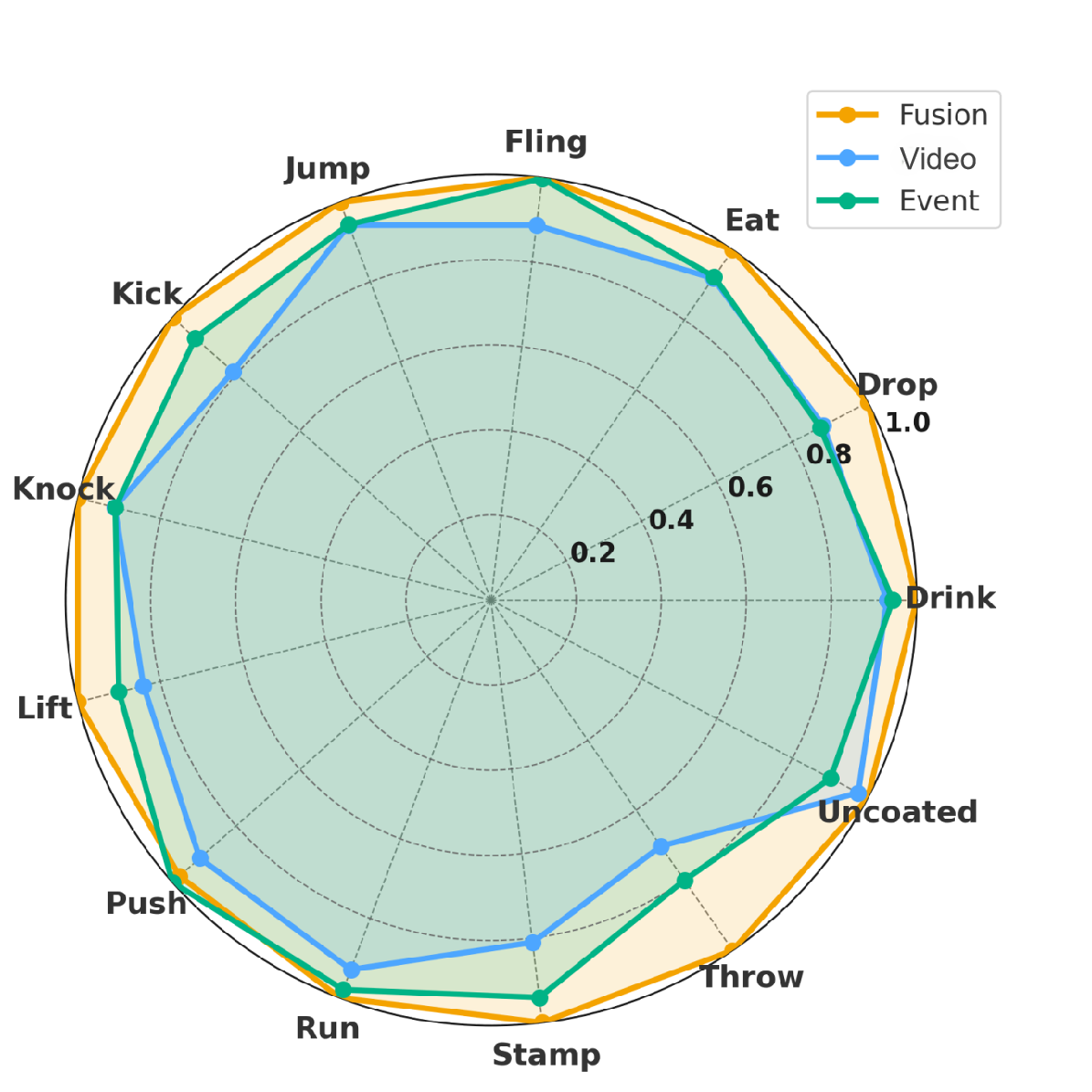}
    \caption{Radar charts showing per-class anomaly detection performance across 13 anomaly categories for image-only, event-only, and fused E-VAD approaches. Each radial axis represents an anomaly class, with values normalized within each category. The chart visualizes the performance distribution of different modalities across diverse abnormal types.}
    \label{fig:radar1}
\end{figure}

\begin{figure}
    \centering
    \includegraphics[width=0.6\linewidth]{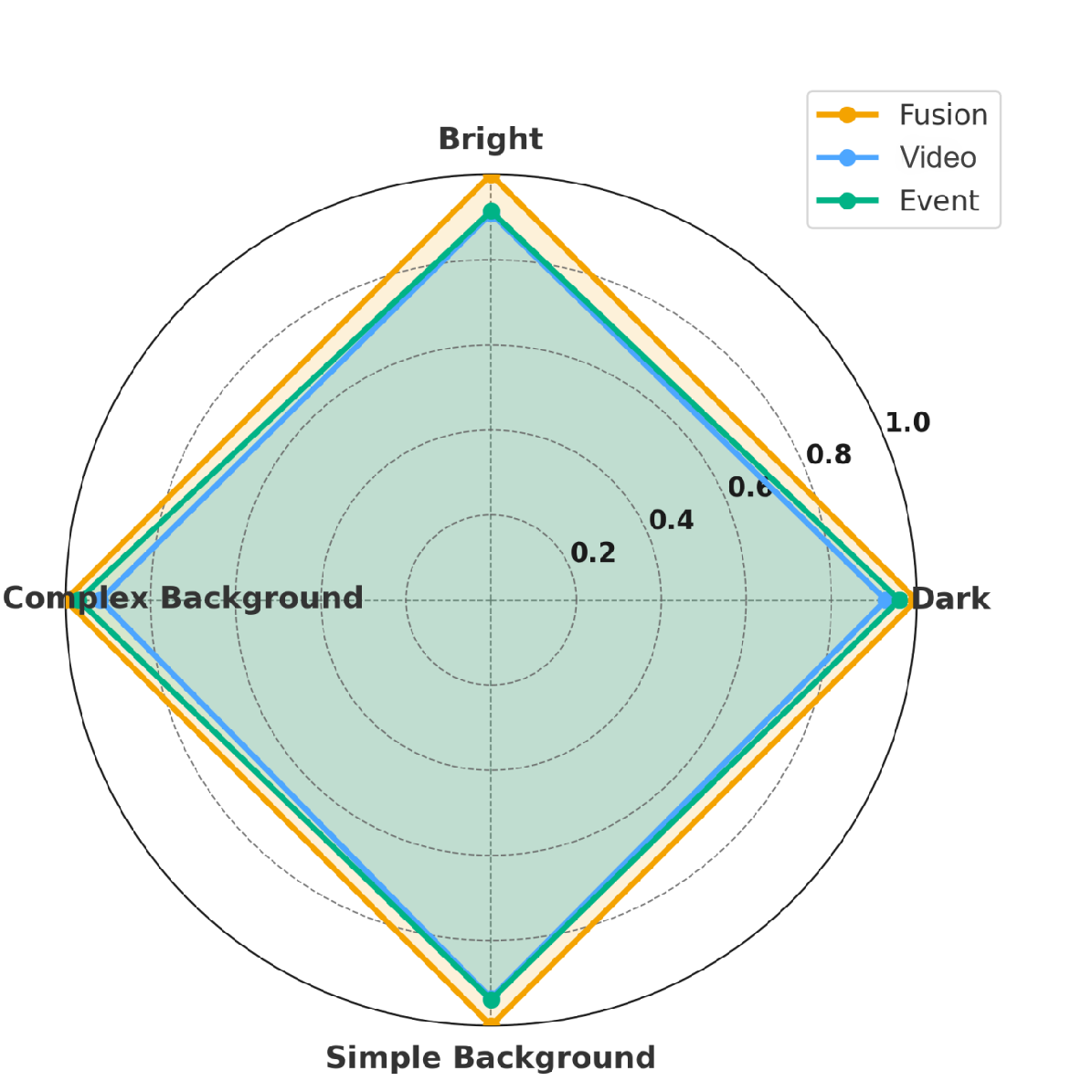}
    \caption{Radar charts showing anomaly detection performance under four environmental conditions for video-only, event-only, and fused approaches. Each radial axis corresponds to a specific condition, and values are normalized per condition. The chart provides a visual comparison of modality behavior across different scene environments.}
    \label{fig:radar2}
\end{figure}

\subsubsection{Influence of Different Event Representations}

We further examine the impact of different event representations on VAD performance, including event frames, voxel grids, and time surfaces. Event frames aggregate events within a fixed temporal interval into a compact 2D representation. For fair comparison, voxel grids are constructed using the same temporal interval and a single temporal bin, resulting in an identical number of events per interval while retaining a raw spatial distribution across the grid. Time surfaces encode the most recent event timestamp at each pixel, producing a continuous representation of local spatiotemporal activity. In all experiments, only event data are used for anomaly detection, without pretraining or multi-modal fusion. As reported in Table \ref{tab:event_representation}, both voxel grids and time surfaces perform substantially worse than event frames. Specifically, voxel grids \cite{gallego2020event} achieve an AUC of 66.40\% with a FAR of 1.00\%, while time surfaces \cite{gallego2020event} reach an AUC of 71.87\% and a FAR of 0.15\%. These results indicate that the performance gap does not stem from reduced temporal coverage or event count, but rather from the representational form itself. Although voxel grids preserve the same event content, they retain a sparse and discontinuous spatial distribution that lacks the accumulated structural patterns formed in event frames. This limits the ability of the model to learn stable and discriminative spatial features under weak supervision. Similarly, time surfaces emphasize recency but provide insufficient spatial structure for robust anomaly discrimination. In contrast, event frames convert recent event activity into compact and discrete motion patterns, yielding more coherent spatial organization and offering a more effective inductive bias for weakly supervised VAD.

\subsection{Qualitative Analysis}
\subsubsection{t-SNE Visualization} To illustrate representation changes after pre-training, we visualize the event, video, and text embeddings using t-SNE, as shown in Fig. \ref{fig:tsne}. In subfigure (a), without multi-modal pre-training, the three modalities exhibit clear domain discrepancies, with scattered clusters and blurred semantic boundaries, indicating that raw event features fail to align consistently with textual semantics and video features. After applying the contrastive multi-modal pre-training, as shown in subfigure (b), the three modalities form compact and well-organized clusters within each semantic category. Event, video, and text features become increasingly aligned within a shared semantic space, demonstrating that the pre-training successfully reduces inter-modal gaps and provides a more discriminative and consistent representation foundation for subsequent anomaly detection.

\subsubsection{Visualization of Detection Results} To illustrate the effectiveness of E-VAD more intuitively, several frame-level detection results are presented in Fig. \ref{fig:vis}. Each example includes video frames, event frames, and the anomaly score curves generated by the video branch, event branch, and fusion branch, with the red shaded region indicating the ground-truth anomaly interval. In dark environments, the video branch struggles to extract discriminative appearance information due to limited visibility, resulting in weak separability between normal and abnormal segments, as illustrated in the dark ``Run", ``Throw", ``Eat", and ``Jump" samples. In contrast, the event branch, less sensitive to illumination changes, maintains relatively stable responses and produces clear peaks during rapid motion. However, when the motion amplitude is small and the duration is long, the inherent sparsity of event signals may lead to discontinuous responses, as shown in the bright ``Lift" sample. The fusion branch, by integrating the strengths of both modalities, yields concentrated and well-defined peaks within the abnormal interval while maintaining low and stable activation in normal frames, thereby aligning more accurately with the ground-truth labels. This demonstrates that multi-modal fusion effectively compensates for the limitations of each single modality and significantly enhances anomaly localization under challenging conditions.

\subsubsection{Results across Classes} 
To further elucidate the advantages of E-VAD in detail, we perform a comparative analysis across distinct abnormal categories using the radar chart in Fig. \ref{fig:radar1}. The chart compares performance of the video branch, event branch, and fusion model across 13 abnormal categories. Radar patterns reveal that the video branch excels in appearance-related categories like uncoated due to its inherent texture and structural representation, yet is vulnerable to motion blur and occlusion, limiting its efficacy in fast-action categories such as throw and stamp. Conversely, the event branch demonstrates robust performance in anomalies with rapid-motion categories (e.g., stamp, fling, push) but shows slightly reduced effectiveness in slow/static categories like uncoated, reflecting its strength in capturing fast motion and weakness in static scenarios. Crucially, the proposed E-VAD fusion method achieves significant improvements across nearly all categories, with an overall envelope substantially larger than single-modality baselines. This underscores how E-VAD effectively harnesses the dynamic strengths of event data and the appearance richness of video data to enable more comprehensive and balanced action discrimination.

\subsubsection{Results across Environments} 
To demonstrate the advantages of E-VAD across diverse environmental conditions, Fig. \ref{fig:radar2} showcases detection performance across four scene scenarios, including bright, dark, complex background, and simple background. The video branch and event branch exhibit similar performance under bright and simple background conditions. However, the video branch experiences notable performance decline in low-light or complex background settings, while the event branch maintains greater stability owing to its illumination-invariant properties and noise-filtering capabilities. The fusion model achieves near-optimal performance across all conditions, underscoring how cross-modal fusion effectively leverages the complementary strengths of event data and visible video imagery to enhance adaptability to environmental variations, thereby enabling robust performance in varied real-world scenarios.

All these qualitative analysis results reveal that E-VAD demonstrates superior robustness and generalization capabilities across a wide range of categories and diverse scene conditions, effectively maintaining consistent performance under varying environmental challenges and category-specific demands.

\section{Conclusion and Future Work}\label{sec:conc}
This work proposes E-VAD, a multi-modal video anomaly detection framework built upon synchronized visible videos and event streams. A new real-world dataset named TJUTCM Pha is introduced as the first event-assisted benchmark for anomaly detection in pharmaceutical laboratory and production environments. Experiments show that event signals provide crucial temporal precision and illumination robustness, while videos offer rich semantic context, and their integration significantly strengthens anomaly perception. The multi-modal pre-training strategy aligns event, video and text features, and the adaptive fusion mechanism enhances temporal responsiveness and semantic consistency, forming a reliable detection paradigm. The results establish a strong foundation for event-driven VAD research and demonstrate the value of leveraging events alongside traditional videos. Future work will explore spiking neural architectures for finer temporal modeling and expand event-driven datasets and foundation models for scalable anomaly understanding.
\bibliographystyle{elsarticle-num}
\bibliography{reference}

\vspace{-1.5cm}

\begin{IEEEbiography}[{\includegraphics[width=1in,height=1.25in,clip,keepaspectratio]{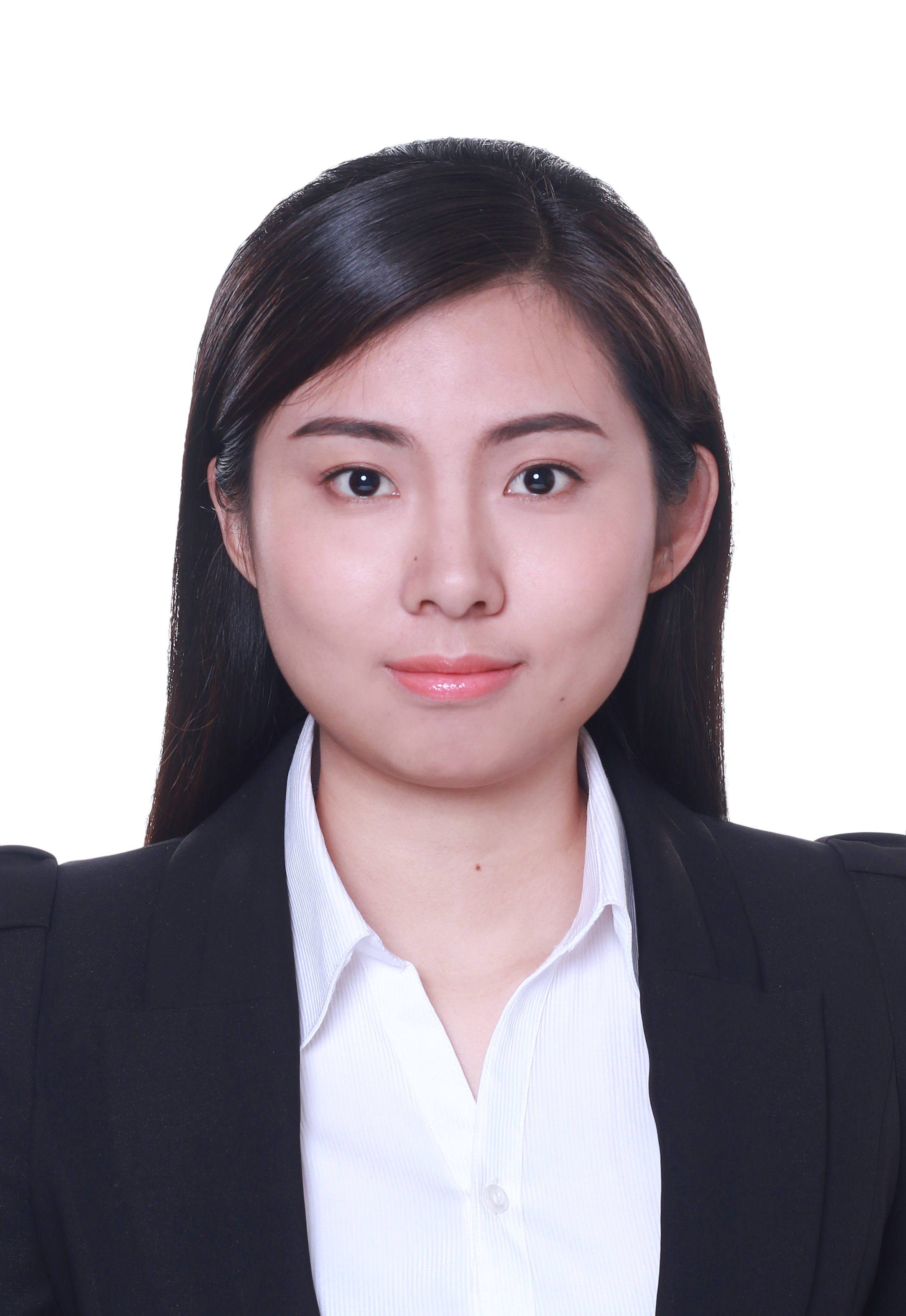}}]{Peipei Zhu} received the B.E. degree from Hebei University of Technology, Tianjin, China, in 2014, the M.E. degree from Beijing University of Posts and Telecommunications, Beijing, China, in 2017, and the Ph.D. degree from the School of Science and Engineering, The Chinese University of Hong Kong (Shenzhen), Shenzhen, China, in 2023. She is currently a Lecturer at the College of Pharmaceutical Engineering of Traditional Chinese Medicine, Tianjin University of Traditional Chinese Medicine. Her current research interests include video processing, intelligent supervision, and embodied intelligence.
\end{IEEEbiography}
\vspace{-1.5cm}

\begin{IEEEbiography}[{\includegraphics[width=1in,height=1.25in,clip,keepaspectratio]{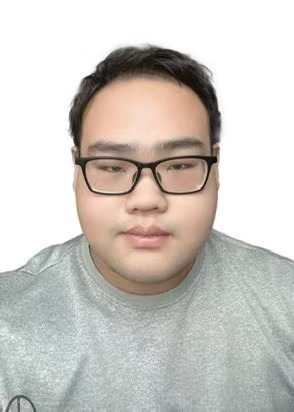}}]{Yueqing Niu}
received the B.E. degree from Tianjin University of Traditional Chinese Medicine, Tianjin, China, in 2024. He is currently a master’s student at the College of Pharmaceutical Engineering of Traditional Chinese Medicine, Tianjin University of Traditional Chinese Medicine.
\end{IEEEbiography}
\vspace{-1.5cm}

\begin{IEEEbiography}[{\includegraphics[width=1in,height=1.25in,clip,keepaspectratio]{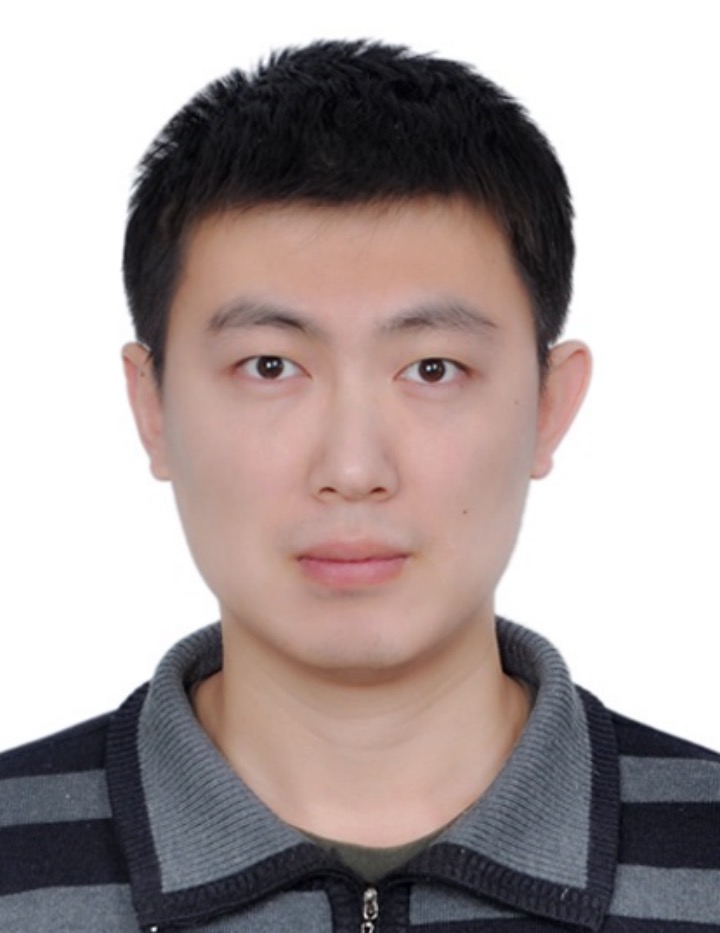}}]{Lin Zhu}
received the B.S. degree from Northwestern Polytechnical University, China, in 2014, the M.S. degree from the North Automatic Control Technology Institute, China, in 2018, and the Ph.D. degree from the School of Computer Science, Peking University, China, in 2022. He is currently an Associate Professor at the School of Artificial Intelligence, Beijing Normal University. His current research interests include AI4Education, video processing, and pattern recognition.
\end{IEEEbiography}
\vspace{-12mm}

\begin{IEEEbiography}[{\includegraphics[width=1in,height=1.25in,clip,keepaspectratio]{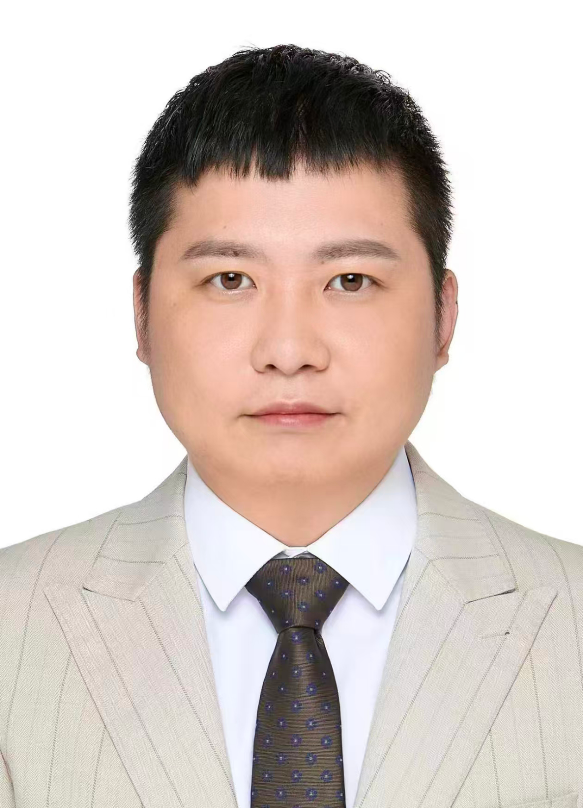}}]{Guanchong Niu} received the B.S. degree in physics from Jilin University, Changchun, China, in 2014, and the Ph.D. degree from The Chinese University of Hong Kong (Shenzhen), Shenzhen, China, in 2021. He was a Senior Engineer at Huawei Technologies Co., Ltd., China, from September 2021 to October 2022. From January to March 2020, he was an Intern at Bell Labs France, where he worked on indoor localization systems. He is currently an Associate Professor at Guangzhou Institute of Technology, Xidian University. His research interests include UAV--UGV cooperative systems, millimeter-wave communications, and MIMO systems for next-generation wireless communications and robotics.
\end{IEEEbiography}
\vspace{-12mm}

\begin{IEEEbiography}[{\includegraphics[width=1in,height=1.25in,clip,keepaspectratio]{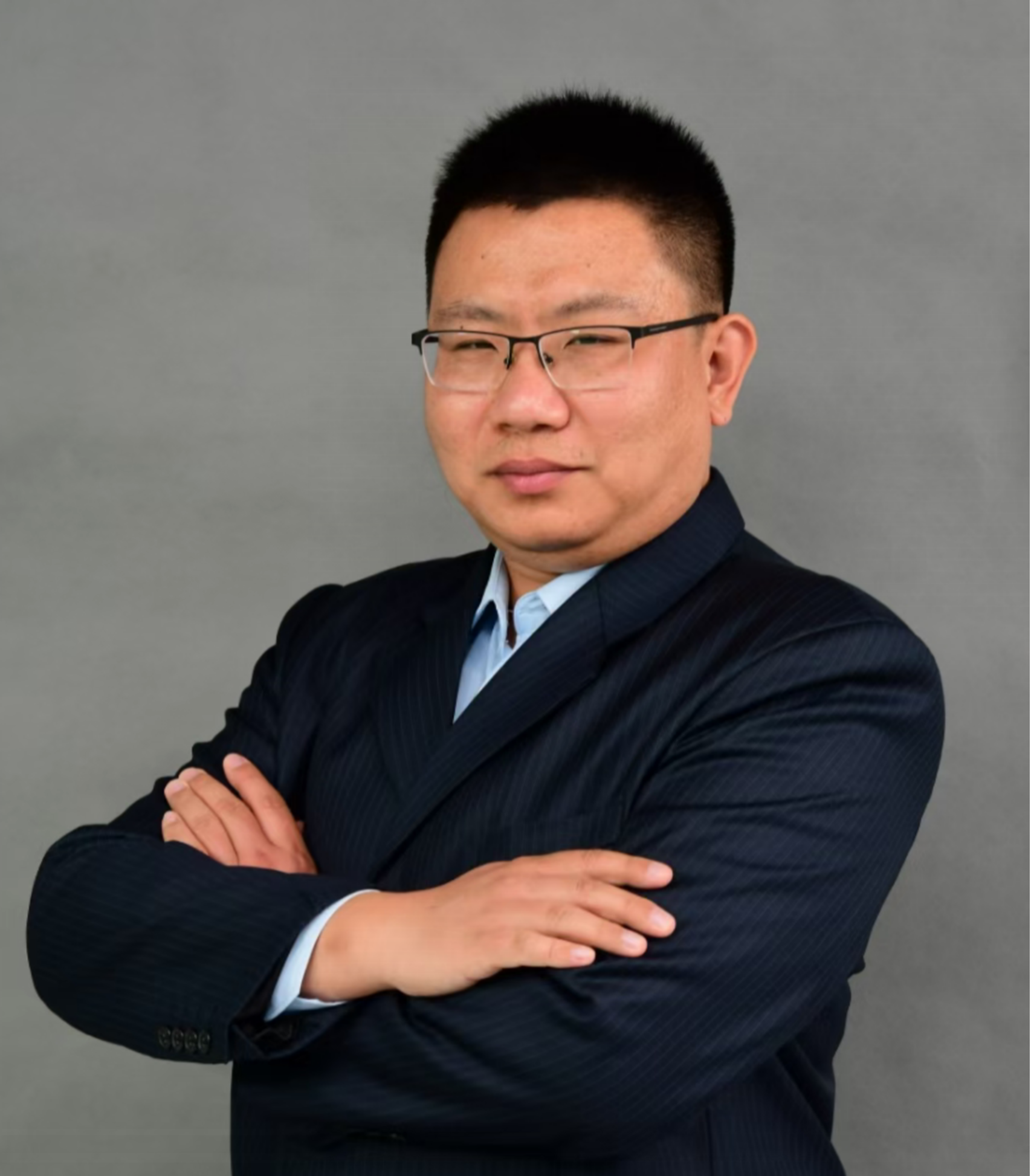}}]{Yang Yu}
received the Ph.D. degree in Chemical Engineering from Tianjin University, Tianjin, China, in 2015. He is currently an Associate Professor at the College of Pharmaceutical Engineering of Traditional Chinese Medicine, Tianjin University of Traditional Chinese Medicine.
\end{IEEEbiography}
\vspace{-12mm}

\begin{IEEEbiography}[{\includegraphics[width=1in,height=1.25in,clip,keepaspectratio]{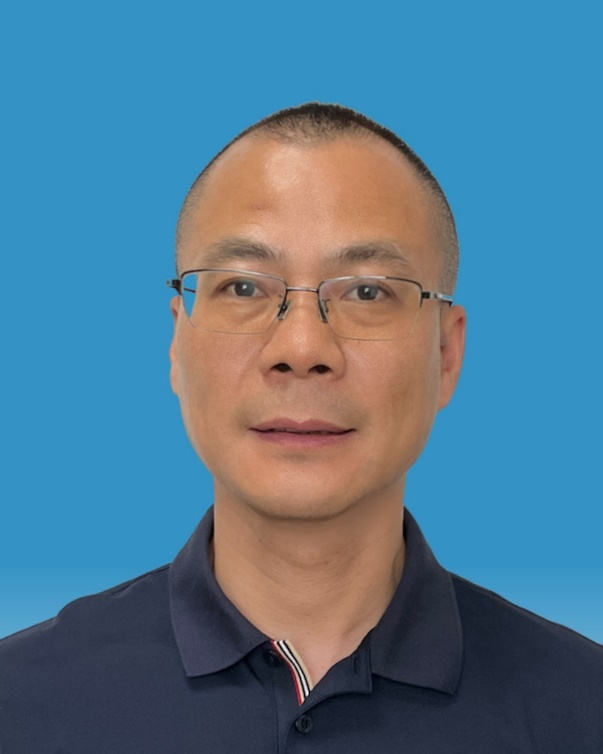}}]{Zheng Li}
received the B.S. and M.S. degrees in Chemical Engineering from Zhejiang University, Hangzhou, China, in 1998 and 2001, respectively, and the Ph.D. degree in Chemical Engineering from Michigan State University, East Lansing, MI, USA, in 2006. He completed his postdoctoral research at Boston University from 2007 to 2008 and worked at Monsanto Company, St. Louis, MO, USA, as a Systems Biology Scientist from 2008 to 2013. He joined Tianjin University of Traditional Chinese Medicine in 2013 as a Full Professor and is currently the Dean of the College of Pharmaceutical Engineering of Traditional Chinese Medicine.
\end{IEEEbiography}

 \vspace{-12mm}
\end{document}